\documentclass[letterpaper, 10 pt, conference]{ieeeconf}  

\IEEEoverridecommandlockouts                              

\usepackage{graphicx} 
\usepackage{amsmath} 
\usepackage{amssymb}  
\usepackage{multirow}

\usepackage{color}
\definecolor{supergood}{rgb}{.2,.6,.2} 
\definecolor{good}{rgb}{.5,.99,.5} 
\definecolor{equal}{rgb}{0.1,0.1,0.1}  
\definecolor{bad}{rgb}{.9,.7,.1} 
\definecolor{superbad}{rgb}{.9,.1,.1}   
\definecolor{notworking}{rgb}{.5,.5,.5}

\usepackage{array}
\newcolumntype{P}[1]{>{\centering\arraybackslash}b{#1}}
\newcommand\VRule[1][\arrayrulewidth]{\vrule width #1}

\usepackage{fancyhdr}
\title{\LARGE \bf
Graph-based non-linear least squares optimization \\ for visual place recognition in changing environments
}

\author{Stefan Schubert, Peer Neubert and Peter Protzel$^{1}$
\thanks{$^{1}$All authors are with Faculty of Electrical Engineering and Automation Technology, Chemnitz University of Technology, 09126 Chemnitz, Germany
        {\tt\small \{firstname.lastname\}@etit.tu-chemnitz.de}}%
}

\begin{document}

\maketitle

\fancyfoot{}
\fancyhead[OL]{ 
    \footnotesize
    To appear in IEEE Robotics and Automation Letters (RA-L), 2021. ACCEPTED VERSION\\
    \tiny
    \copyright 2020 IEEE. Personal use of this material is permitted.  Permission from IEEE must be obtained for all other uses, in any current or future media, including reprinting/republishing this material for advertising or promotional purposes, creating new collective works, for resale or redistribution to servers or lists, or reuse of any copyrighted component of this work in other works.
}
\addtolength{\headheight}{\baselineskip}
\thispagestyle{fancy}
\pagestyle{empty}

\begin{abstract}
Visual place recognition is an important subproblem of mobile robot localization.
Since it is a special case of image retrieval, the basic source of information is the pairwise similarity of image descriptors.
However, the embedding of the image retrieval problem in this robotic task provides additional structure that can be exploited, e.g. spatio-temporal consistency.
Several algorithms exist to exploit this structure, e.g., sequence processing approaches or descriptor standardization approaches for changing environments.
In this paper, we propose a graph-based framework to systematically exploit different types of additional structure and information. 
The graphical model is used to formulate a non-linear least squares problem that can be optimized with standard tools.
Beyond sequences and standardization, we propose the usage of intra-set similarities within the database and/or the query image set as additional source of information.
If available, our approach also allows to seamlessly integrate additional knowledge about poses of database images.
We evaluate the system on a variety of standard place recognition datasets and demonstrate performance improvements for a large number of different configurations including different sources of information, different types of constraints, and online or offline place recognition setups.

\end{abstract}

\section{INTRODUCTION}
\begin{figure}[tb]
 \centering 
 \includegraphics[width=1\linewidth]{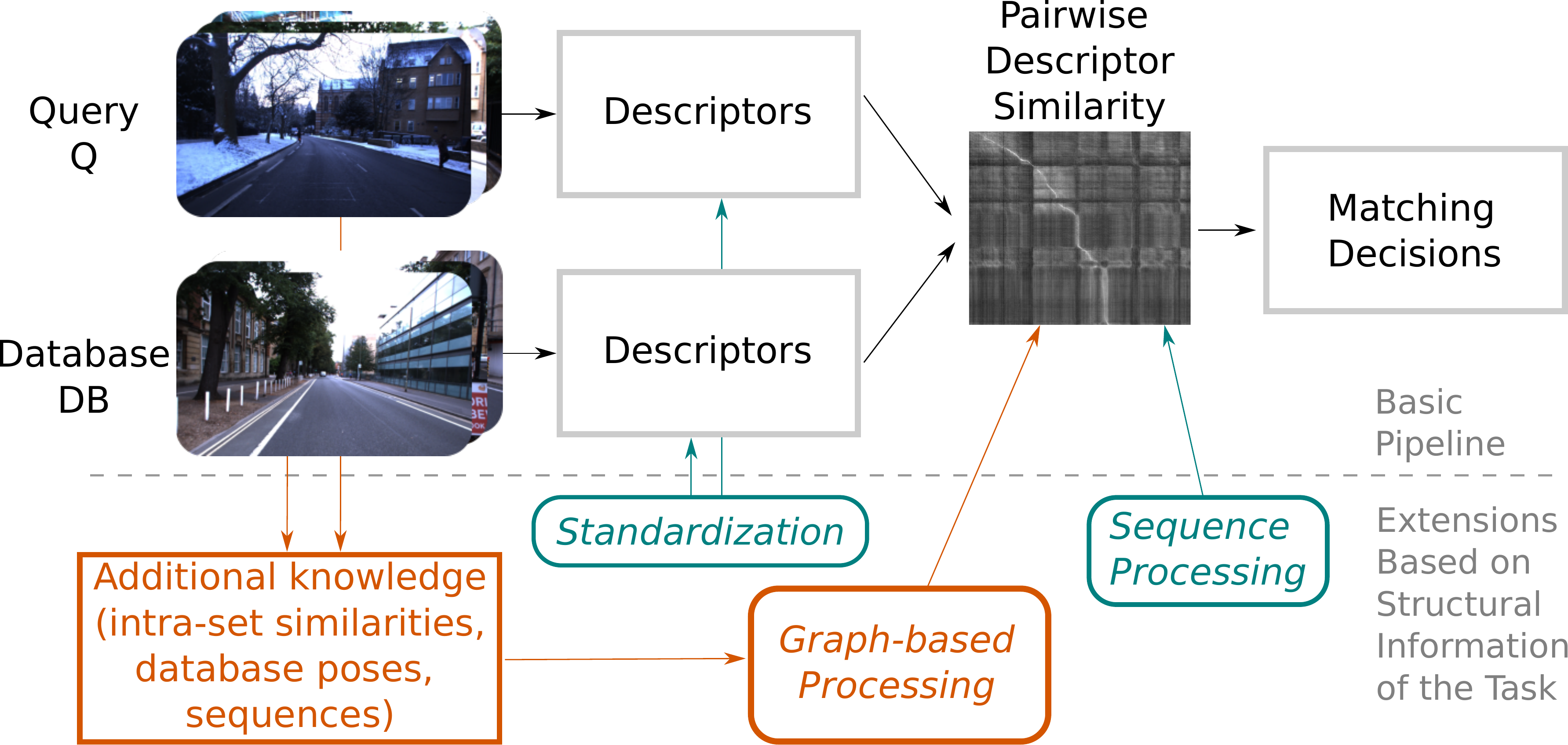}
 \vspace{-0.7cm}
 \caption{The basic place recognition pipeline (above the horizontal dashed line) can be extended with additional information (below this line). Established approaches are standardization of descriptors and sequence processing. We propose a graph-based framework to integrate various sources of additional information in a common optimization formulation. An example for so far rarely used information are intra-set similarities.} 
 \vspace{-0.4cm}
 \label{fig:intro}
\end{figure}

Visual place recognition in changing environments is the problem of finding matchings between two sets of observations, database and query, despite severe appearance changes.
It is required for loop closure detection in SLAM (Simultaneous Localization and Mapping) and for candidate selection in image-based 6D localization systems.
The common pipeline for visual place recognition in changing environments is shown in Fig.~\ref{fig:intro} (top):
Given the two sets of images, database (the ``map'') and query (later or current run), a descriptor is computed for each image.
Subsequently, each descriptor of the database is compared pairwise to each descriptor of the query to get a similarity matrix $\hat{S}^{DB\times Q}$ that can finally be used to determine potential place matches.
As illustrated in Fig.~\ref{fig:intro} (bottom), there are various additional approaches to improve the place recognition pipeline either by preprocessing the descriptors, e.g., with feature standardization, or by postprocessing the similarity matrix, e.g., with sequence search in the similarity matrix.
In addition to these pre- and postprocessing steps, additional information can be exploited like database image poses or the so far rarely used intra-database and intra-query similarities.
However, there are only few methods that exploit such additional knowledge in order to further improve the place recognition performance.

In this paper, we address the problem of systematically exploiting additional information by proposing a versatile, expandable framework for different kinds of prior knowledge from or about database and query.
Specifically, we
\begin{itemize}
 \item propose a versatile, expandable graph-based framework that formulates place recognition as a sparse non-linear least squares optimization problem in a factor graph (Sec.~\ref{sec:algo})
 \item discuss different sources of information in place recognition problems and propose a loop-rule and an exclusion-rule to exploit inherent structural properties of the place recognition problem (Sec.~\ref{sec:binary_factor})
 \item demonstrate how this framework can be used to integrate the different sources of information, e.g., we provide implementations of the loop- and exclusion-rules in terms of cost functions for factors. These either exploit intra-set descriptor similarities within database and/or query, or, if available, additional knowledge about database image poses (Sec.~\ref{sec:binary_factor})
 \item demonstrate the versatility of the framework by proposing an n-ary factor that mimics the sequence processing approach of SeqSLAM \cite{seqSLAM} (Sec.~\ref{sec:nary_factor})
 \item present the optimization using standard non-linear least squares optimization tools (Sec.~\ref{sec:optim})
 \item provide comprehensive experimental evaluation on a variety of datasets, configurations and existing methods from the literature (Sec.~\ref{sec:results})
\end{itemize}

The paper closes with a discussion of the current implementation and potential extensions in Sec.~\ref{sec:discussion}.

\section{RELATED WORK}
\label{sec:related_work}
Visual place recognition in changing environments is a subject of active research.
An overview of existing methods is given in \cite{Lowry2016}.
In the present paper, the basic source of information to match places are image descriptors.
The authors of \cite{Suenderhauf2015} demonstrated that intermediate CNN-layer outputs like the \textit{conv3}-layer from AlexNet \cite{alexnet} trained for image classification can serve as holistic image descriptors to match places across condition changes between database and query.
Moreover, there are CNNs especially trained for place recognition that return either holistic image descriptors like NetVLAD \cite{netvlad} or local features like DELF \cite{delf}.
The performance of holistic descriptors can be further improved by additional pre- and postprocessing steps.
To improve the performance of holistic descriptors, feature standardization and unsupervised learning techniques like PCA- and clustering-based methods can be used \cite{Schubert2020}.
In \cite{Neubert2019} it is shown how a neurologically-inspired neural network can be used to combine several descriptors along a sequence to a new descriptor for each place.
\cite{Schubert2019} extended this approach to encode additional odometry information in the new descriptors.
Given the final pairwise similarities between descriptors from database and query, sequence-based methods \cite{seqSLAM}\cite{hmm} for postprocessing can be used to improve the place recognition performance further.

In this paper, we propose a graph-based approach to optimize the descriptor similarities by incorporating prior knowledge.
In \cite{hmm} Hansen and Browning used a hidden Markov model (HMM) to formulate a graph-based sequence search method in the similarity matrix.
Naseer et al. \cite{Naseer14} used a flow network with edges defined for sequence search to improve place recognition results.
In \cite{Zhang2019} a graph is used to prevent place matches between adjacent places and to distribute high matching scores to neighboring places.
In contrast to these approaches, our approach exploits not only sequence information but also integrates other additional knowledge, e.g., about intra-set similarities in the database set or the query set.
The literature provides several approaches for localization where the database is known in advance.
For example, given the images from the database (or another representative training set), FabMap \cite{fabmap} learns statistics about feature occurrences in order to determine the most descriptive features.
McManus et al. \cite{McManus2014} train offline condition-invariant broad-region detectors from beforehand collected images with a variety of appearances at particular locations.
In \cite{Neubert2015} intra-database descriptor comparisons are used to reduce the number of required image comparisons during the query run.
Vysotska and Stachniss \cite{Vysotska2017} use binary intra-database place matches to perform jumps within the similarity matrix during sequence search.
In contrast, in our presented approach we use potentially continuous intra-database similarities to optimize the place recognition result instead of just accelerating it.
Moreover, we do not only use this information to find loops but in addition to potentially inhibit wrong loop closures.

Graphical models, and in particular factor-graphs, are a well established tool in mobile robotics \cite{Dellaert2017}, e.g., in the form of robust pose graph SLAM \cite{Sunderhauf2012}.
Similar to the here proposed approach, pose graph SLAM represents each place as a node and the edges (factors) model constraints between these places. 
However, a significant difference is that pose graph SLAM deals with spatial information, i.e., the places are represented by pose coordinates in the world and the factors are spatial transformations between these poses (e.g., odometry or detected loop closures). 
In contrast, our presented approach is intended to be used \textit{before} SLAM to establish loop closures. 
In particular, we do not optimize metric errors between spatial constraints but errors in the mutual consistency of descriptor similarities.

\section{ARCHITECTURE OF THE GRAPHICAL MODEL}\label{sec:algo}
A graphical model serves as a structured representation of prior knowledge in terms of dependencies, rules, and available information.
Given the variable nodes in the graph with their initial values together with dependencies between nodes based on additional knowledge, optimization algorithms can be used to modify the variables in order to resolve inconsistencies between nodes.
We are going to exploit this mechanism and present a graph-based framework for visual place recognition that optimizes the initial similarity values $\hat{S}^{DB\times Q}$ from pairwise image descriptor comparisons.

The graph-based framework is expressed as factor graph.
Factor graphs are graphical representations of least squares problems -- for a detailed introduction to factor graphs please refer to \cite{Dellaert2017}.
The graph's basic architecture consists of nodes with unary factors that penalize deviations from the initial similarity values.
Depending on additional knowledge, we can add different factors in the graph to introduce connections (i.e. dependencies) between nodes.
Two architectures with nodes, unary factors, and additional factors are shown in Fig.~\ref{fig:graph_intra_set}. 
Each factor defines a quadratic cost function based on the variables it connects. 
The resulting combined optimization problem is defined as a weighted sum of the individual cost functions.
The optimization is subject of Sec.~\ref{sec:optim}.
Here, we first define the basic architecture of the graph with corresponding nodes and unary factors.
Next, we propose two ways to exploit prior knowledge with binary and n-ary factors.
We will structure the explanation of each factor by the \textit{prior knowledge} that can be exploited, a corresponding \textit{rule} that expresses the resulting dependency between nodes, a proposed related \textit{cost function} $f(\cdot)$ that punishes a violation of this rule, and the factor's \textit{usage} in the graph.
Except for the unary factor, each factor is optional and depends on the available knowledge.
Accordingly, the proposed framework can be extended in future work with additional factors.

\subsection{Graph nodes}\label{sec:algo:nodes}
The graph-based framework is designed to optimize the initial pairwise descriptor similaritiy matrix $\hat{S}^{DB\times Q} \in \mathbb{R}^{M\times N}$.
$M$ is the number of database images and $N$ the number of query images.
Accordingly, we define $M\times N$ nodes where each node $s_{ij} \in S^{DB\times Q}$ is a variable version of its initial value $\hat{s}_{ij} \in \hat{S}^{DB\times Q}$ with $s_{ij},\hat{s}_{ij} \in [0,1]$.

\subsection{Unary factor} \label{sec:unary_factor}
\subsubsection{Prior Knowledge}
Descriptor similarities $\hat{s}_{ij}$ are the primary source of information for place recognition.
Beyond initializing the variables to these similarities, we have to prevent too large deviations of $s_{ij}$ from $\hat{s}_{ij}$, in particular to prevent trivial solutions during optimization.

\subsubsection{Rule ''prior``}
\begin{align}\label{eq:rule_unary}
 s_{ij} \approx \hat{s}_{ij}
\end{align}

\subsubsection{Cost function ''prior``}
\begin{align}\label{eq:cost_unary}
 f_\text{prior}(\cdot) = (s_{ij} - \hat{s}_{ij})^2
\end{align}

\subsubsection{Usage}
Each node $s_{ij}$ is connected with a single unary factor to its initial value $\hat{s}_{ij}$ as shown in Fig.~\ref{fig:graph_intra_set}.
Thus, $M\times N$ unary factors are used in a graph.

\subsection{Binary factors for the exploitation of intra-database or intra-query similarities from poses or descriptor-comparisons}\label{sec:binary_factor}

\subsubsection{Prior Knowledge -- intra-database similarities from poses}
In some applications, the database is created with a more advanced sensor setup than the query.
Due to the missing position information for the query images, direct position-based matching cannot be conducted.
Nevertheless, available poses for database images can be used to create binary intra-database similarities $\hat{s}^{DB}_{ij} \in \hat{S}^{DB}$ that encode whether two database images $i$ and $j$ show the same place ($\hat{s}^{DB}_{ij}:=1$) or different places ($\hat{s}^{DB}_{ij}:=0$).

\subsubsection{Prior Knowledge -- intra-set similarities from image descriptors}
\begin{figure}[tb]
 \centering
 \includegraphics[width=0.8\linewidth]{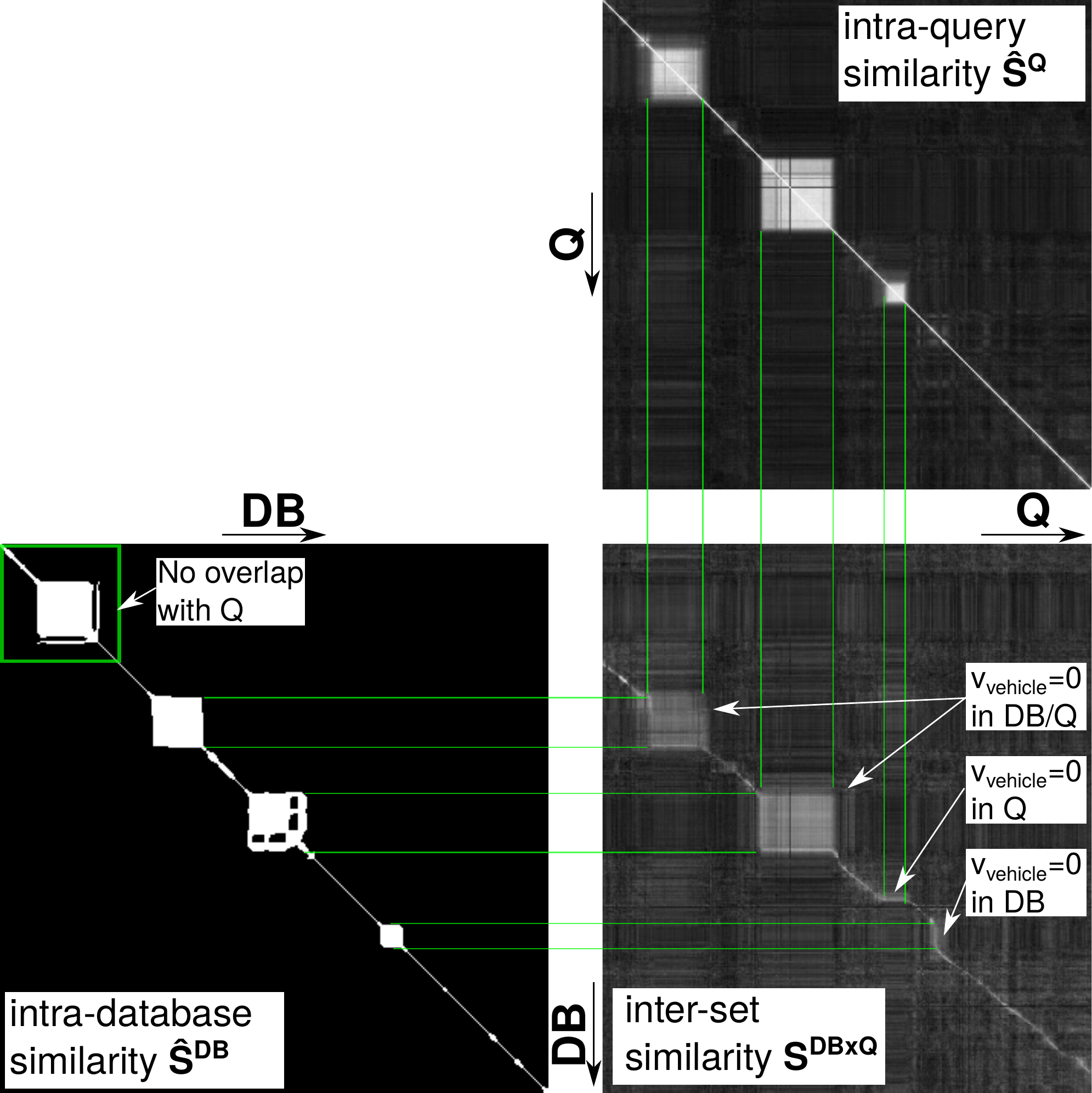}
 \vspace{-0.2cm}
 \caption{
 The information of the similarity matrix between database and query (bottom right) can be extended with similarity information of images \textit{within} the query set (top right) and/or \textit{within} the database set (bottom left).}
 \vspace{-0.2cm}
 \label{fig:s_matrices}
\end{figure}

Even if position data is not available, we can acquire similar information directly from image comparisons within the database and also within the query to get intra-database similarities $\hat{s}^{DB}_{ij} \in \hat{S}^{DB}$ and intra-query similarities $\hat{s}^Q_{ij} \in \hat{S}^Q$ with $\hat{s}^{DB}_{ij}, \hat{s}^Q_{ij} \in [0,1]$.
These intra-set similarities are an inherent and almost always available source of additional information, which has not been used often yet.
The computation of intra-set similarities can usually be done more reliably than the computation of inter-set similarities, because the condition \textit{within} database or query is potentially more constant than between both. 
The intra-set image comparisons could be done with methods like FabMap~\cite{fabmap} that are more suited for place recognition under constant condition.
Fig.~\ref{fig:s_matrices} shows how the intra-set similarities $\hat{S}^{DB}$, $\hat{S}^Q$ are related to the inter-set similarities $\hat{S}^{DB\times Q}$ and how they can reveal loops and zero-velocity stages.

We define separate binary factors for both intra-set similarities, either from poses or from image descriptors.
Each binary factor is a combination of two complementing rules and cost functions $f^Q_\text{loop}(\cdot)+f^Q_\text{exclusion}(\cdot)$ (Eq.~(\ref{eq:cost_sim_q}), (\ref{eq:cost_dissim_q})) and $f^{DB}_\text{loop}(\cdot)+f^{DB}_\text{exclusion}(\cdot)$ (Eq.~(\ref{eq:cost_sim_db}), (\ref{eq:cost_dissim_db})), respectively.

\subsubsection{Rule ''loop``}
If $\hat{s}^{Q}_{jl}$ is high (indicated by ''$\hat{s}^{Q}_{jl}\uparrow$``), the $j$-th and $l$-th query image likely show the same place.
If the $i$-th database image is compared to these two query images, the following ternary relation can be derived:
\begin{align}\label{eq:rule_sim2}
    s_{ij} \approx s_{il} &\text{, iff } \hat{s}^Q_{jl}\uparrow
\end{align}

The rule expresses that ''$s_{ij}$ should be equal to $s_{il}$ iff the query images $j$, $l$ show the same place``, because if both query images $j$, $l$ show the same place, database image $i$ can either be equal to both or to none of both.
Accordingly, this rule exploits loops within the intra-query similarities.
This rule is inherent to the place recognition problem and valid as well for intra-database similarities:
\begin{align}\label{eq:rule_sim1}
    s_{ij} \approx s_{kj} &\text{, iff } \hat{s}^{DB}_{ik}\uparrow
\end{align}

\subsubsection{Cost function ''loop``}
For equation (\ref{eq:rule_sim2}) and (\ref{eq:rule_sim1}), cost functions similar to (\ref{eq:cost_unary}) can be formulated:
\begin{align}
 f^Q_\text{loop}(\cdot) &= \hat{s}^Q_{jl} \cdot (s_{ij} - s_{il})^2 \label{eq:cost_sim_q} \\
 f^{DB}_\text{loop}(\cdot) &= \hat{s}^{DB}_{ik} \cdot (s_{ij} - s_{kj})^2 \label{eq:cost_sim_db} 
\end{align}
The quadratic error term is multiplied by the intra-set similarity $\hat{s}^Q_{jl}$ or $\hat{s}^{DB}_{ik}$ to apply weighting in case of non-binary intra-set similarities from image descriptors or to turn it on and off for binary intra-set similarities.

\subsubsection{Rule ''exclusion``}
If $\hat{s}^{Q}_{jl}$ is low (indicated by ''$\hat{s}^{Q}_{jl}\downarrow$``), the $j$-th and $l$-th query image probably show different places.
If the $i$-th database image is compared to these two query images, the following ternary relation can be derived:
\begin{align}\label{eq:rule_dissim2}
    \neg(s_{ij}\uparrow \land s_{il}\uparrow) 
    &\text{, iff } \hat{s}^Q_{jl}\downarrow
\end{align}
This rule expresses that ''not both similarity measurements $s_{ij}$ AND $s_{il}$ can be high iff the query images $j$, $l$ show different places``, i.e., the rule excludes one or both similarities $s_{ij}$, $s_{il}$ from being high; otherwise a single database image $i$ would show two different places concurrently.
Again, this rule is inherent to the place recognition problem and is supposed to add valuable information that can be exploited.
As before, the same rule is valid for intra-database similarities:
\begin{align}\label{eq:rule_dissim1}
    \neg(s_{ij}\uparrow \land s_{kj}\uparrow) 
    &\text{, iff } \hat{s}^{DB}_{ik}\downarrow
\end{align}

\subsubsection{Cost function ''exclusion``}
It is less natural how to express the \textit{rule ''exclusion``} in a cost function. This is part of the discussion in Sec.~\ref{sec:discussion}.
In this work, we define the following cost functions for equation (\ref{eq:rule_dissim2}) and (\ref{eq:rule_dissim1}):
\begin{align}
 f^Q_\text{exclusion}(\cdot) &= (1-\hat{s}^Q_{jl}) \cdot (s_{ij} \cdot s_{il})^2 \label{eq:cost_dissim_q} \\
 f^{DB}_\text{exclusion}(\cdot) &= (1-\hat{s}^{DB}_{ik}) \cdot (s_{ij} \cdot s_{kj})^2 \label{eq:cost_dissim_db}
\end{align}
The quadratic error term is multiplied by the negated intra-set similarity $(1-\hat{s}^Q_{jl})$ or $(1-\hat{s}^{DB}_{ik})$ to weight it in case of non-binary intra-set similarities from image descriptors or to turn it on and off for binary intra-set correspondences.

\subsubsection{Usage}
\begin{figure}[tb]
 \centering
 \includegraphics[width=\linewidth]{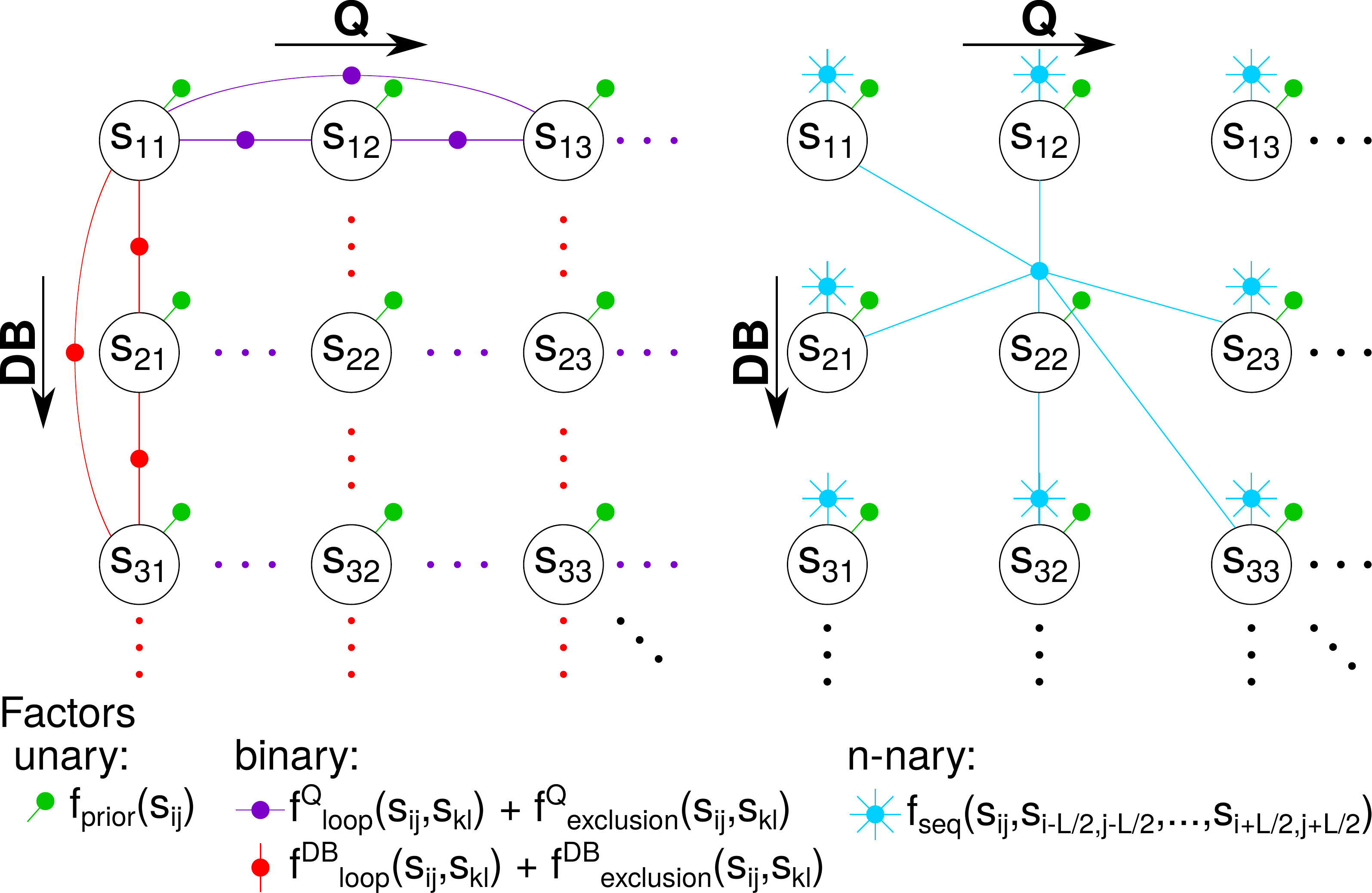}
 \caption{Illustration of the graph structure. \textit{(left)} Unary and binary factors. \textit{(right)} Unary and n-nary factors, which connect structured local blocks.}
 \label{fig:graph_intra_set}
\end{figure}

Fig.~\ref{fig:graph_intra_set} (left) shows a graphical model with nodes $s_{ij}$, unary factors $f_\text{prior}(\cdot)$, and the proposed binary factors.
To apply the binary factors for existing intra-database similarities $\hat{S}^{DB}$, each node $s_{ij}$ has to be connected to every node $s_{kj}$ for all $k = 1,\ldots,M$ with $k\neq i$ within each column in $S^{DB\times Q}$; i.e., $\binom{M}{2}$ factors per column.
For existing intra-query similarities $\hat{S}^Q$, each node $s_{ij}$ has to be connected to every node $s_{il}$ for all $l = 1,\ldots,N$ with $l\neq j$ within each row in $S^{DB\times Q}$; i.e., $\binom{N}{2}$ factors per row.
The potentially high number of connections is part of the discussion in Sec.~\ref{sec:discussion}.

\subsection{N-ary factors for the exploitation of sequences}\label{sec:nary_factor}

\subsubsection{Prior Knowledge}
If both database and query are recorded as spatio-temporal sequence, sequences also appear in the inter-set similarities $S^{DB\times Q}$ (see Fig.~\ref{fig:s_matrices}).
In this case, SeqSLAM \cite{seqSLAM} showed the benefit from a simple combination of similarities of neighboring images.
Originally, it was implemented as postprocessing of the similarity matrix $S^{DB\times Q}$. 
We can integrate such sequential information within the graph in a similar fashion using n-ary factors.

\subsubsection{Rule ''sequence``}
\begin{align}\label{eq:rule_seq}
 s_{ij} \approx \frac{1}{L}\sum_{\forall \{k,l\} \in \text{Seq}({i,j};L,v_p)} s_{kl}
\end{align}
$\text{Seq}({i,j};L,v_p)$ is a function that returns similarities that are part of a sequence along a line segment with slope $v_p$ and sequence length $L$ within $S^{DB\times Q}$ around $s_{ij}$.
For a full explanation how SeqSLAM works, please refer to \cite{seqSLAM}.

\subsubsection{Cost function ''sequence``}

\begin{align}
  f_\text{seq}(\cdot) = (s_{ij} - \frac{1}{L}\max_{v_p \in V}{(\sum_{\{k,l\} \in \text{Seq}(\{i,j\};L,v_p)} s_{kl})})^2
\end{align}
with $V$ being the set of allowed velocities within $S^{DB\times Q}$.

\subsubsection{Usage}
Fig.~\ref{fig:graph_intra_set} (right) shows a graphical model with nodes $s_{ij}$, unary factors $f_\text{prior}(\cdot)$, and the proposed n-ary factors for sequences.
As for the unary factors in Sec.~\ref{sec:unary_factor}, each node in the graph is equipped with one n-ary factor.
Each n-ary factor connects its node to all nodes that are part of any sequence with length $L$ and slope $v_p\in V$.
Accordingly, $M\times N$ n-ary factors are introduced into a graph if sequences are exploited.

\section{OPTIMIZATION OF THE GRAPHICAL MODEL}\label{sec:optim}
The objective of the graph-based framework is the formulation of dependencies from prior knowledge for a subsequent optimization of the similarities $s_{ij}$ to incorporate the prior knowledge into the final similarity values, and to potentially resolve contradictory dependencies.

We use factor graphs as graphical representation of least squares problems and defined every cost function for each type of factor as (weighted) quadratic error function.
Accordingly, the global error $E$ is defined as a sum over the (weighted) cost functions of all factors in the graph.

\subsection{Weighting of factor's costs for global error computation in the graph}
As usually done in error computation for a graph, costs from different factor-types have to be balanced by weighting them separately.
Therefore, we normalize the cost of each factor by the number of factors per factor-type, and introduce user-specified weights $w$ for all factors except for the basic unary factor (which gets weight 1).

For the unary factor of our basic architecture with cost function $f_\text{prior}(\cdot)$, we define the partial global error $E_\text{prior}$ with
\begin{align}
 E_\text{prior} = \frac{1}{MN} \sum_{i=1}^M \sum_{j=1}^N f_\text{prior}(s_{ij})
\end{align}

In case of dependencies in the graph that are introduced by available intra-database or intra-query similarities, we define a partial global error $E^{DB}_{loop,exclusion}$ or $E^Q_{loop,exclusion}$ for the weighted summation over all binary factors (Sec.~\ref{sec:binary_factor}): 
\begin{align}
 E^{DB}_\text{loop,exclusion} = &\frac{1}{N\binom{M}{2}} \sum_{j=1}^N \sum_{i=1}^{M-1} \sum_{k=i+1}^M w^{DB}_\text{loop} \cdot f^{DB}_\text{loop}(s_{ij}, s_{kj}) \nonumber\\
 &+ w^{DB}_\text{exclusion} \cdot f^{DB}_\text{exclusion}(s_{ij}, s_{kj})
 \\
 E^Q_\text{loop,exclusion} = &\frac{1}{M\binom{N}{2}} \sum_{i=1}^M \sum_{j=1}^{N-1} \sum_{l=j+1}^N w^Q_\text{loop} \cdot f^Q_\text{loop}(s_{ij}, s_{il}) \nonumber\\
 &+ w^Q_\text{exclusion} \cdot f^Q_\text{exclusion}(s_{ij}, s_{il})
\end{align}

For the n-ary factors (Sec.~\ref{sec:nary_factor}) in case of sequence exploitation, the partial global error $E_\text{seq}$ is defined with
\begin{align}
 E_\text{seq} = \frac{1}{MN} \sum_{i=1}^M \sum_{j=1}^N w_\text{seq} \cdot f_\text{seq}(s_{ij}, \text{Seq}(\{i,j\};L,v^*_{ij})
\end{align}
with $v^*_{ij}$ being the optimal velocity (slope) with the highest mean of connected similarities around $s_{ij}$.

Finally, summation over all partial global errors $E_i$ that occur in the graph yields the global error $E$:
\begin{align}
 E = \sum_{\forall i} E_i
\end{align}

\subsection{Implementation of the optimization}
Error $E$ is a sum solely over quadratic cost functions.
Thus, many tools for non-linear least squares (NLSQ) optimization can be used (e.g., scipy's \textit{least\_squares}-function in Python).
For an easier formulation of the optimization problem, factor graph tools like g2o \cite{g2o} can be used.
However, one should be aware that depending on the introduced factors, the optimization problem can get quite huge.
Thus, efficient implementations should be preferred that perform a quick and memory efficient optimization.
Sec.~\ref{sec:seq} reports our achieved runtimes that are presumably sufficient for many applications.
Sec.~\ref{sec:discussion} provides some more discussion on alternative optimization methods and approximation techniques for runtime improvements.

\section{EXPERIMENTAL RESULTS}\label{sec:results}
We present experiments that investigate the performance gains achieved by the graph-based framework with three different extensions that exploit 1)~intra-database similarities; 2)~intra-database and intra-query similarities; 3)~intra-database similarities, intra-query similarities and sequences.
In order to evaluate the potential benefit beyond available pre- and postprocessing methods, we repeat these experiments with the descriptor standardization approach from \cite{Schubert2020} for preprocessing and SeqSLAM for postprocessing.
In a final experiment, we compare our graph-based method with sequence-based approaches from the literature.

\subsection{Experimental Setup}
\subsubsection{Image descriptor}
NetVLAD \cite{netvlad} is used as CNN-image descriptor in all experiments.
We use the author's implementation trained on the Pitts30k dataset with VGG-16 and whitening.

\subsubsection{Metric}
The performance is measured with average precision which is the area under the precision-recall curve.

\subsubsection{Datasets}
All experiments are based on the five different datasets Nordland~\cite{nordland}, StLucia (Various Times of the Day)~\cite{stlucia}, CMU (Visual Localization)~\cite{cmu}, GardensPoint Walking~\cite{gardenspoint} and Oxford RobotCar~\cite{robotcar} with different characteristics regarding the environment, appearance changes, in-sequence loops, stops, or viewpoint changes.
We use the datasets as described in our previous work~\cite{Neubert2019}.
Images from StLucia, CMU and Oxford were sampled with one frame per second, which preserves varying camera speeds, stops and loops in the datasets, and leads to translation and orientation changes during revisits. GardensPoint contains lateral viewpoint changes.

For the binary intra-database similarities $\hat{S}^{DB}_{pose}$ from poses, we use the GPS from the datasets or a main diagonal for \textit{GardensPoint Walking} and \textit{Nordland}.
For the query sequence, we assume that no GPS is available.

\subsubsection{Implementation}\label{sec:implementation}
Graph creation and optimization were implemented in Python with scipy's \textit{least\_squares}-optimization function; the \textit{Trust Region Reflective} algorithm was used for minimization.
Due to the huge number of factors within a graph in case of the usage of intra-database and intra-query similarities, we divided $S^{DB\times Q}$ into equally sized patches with height and width $\leq 500$, and optimized each patch separately.
No information is shared between patches, and the n-ary factors are truncated at borders.
A full optimization on $S^{DB\times Q}$ without patches was performed if only intra-database similarities $\hat{S}^{DB}$ were used.
The variables in the optimization are initialized with $\hat{S}^{DB\times Q}$ from the pairwise descriptor comparisons.

\subsubsection{Parameters}
In all experiments, we used a fixed parameter set that was determined from a toy-example and a small real-world dataset.
We used $w^{DB}_\text{loop}=4, w^{DB}_\text{exclusion}=40$ for intra-set similarities from GPS, $w^{DB}_\text{loop}=w^Q_\text{loop}=1, w^{DB}_\text{exclusion}=w^Q_\text{exclusion}=20$ for intra-set similarities from descriptors, and $w_\text{seq}=10$, $L=11$ for sequences.

\subsection{Contributions of information sources and rules}\label{sec:exp1}
In Sec.~\ref{sec:algo} we identified four rules: \textit{''prior``}, \textit{''loop``}, \textit{''exclusion``} and \textit{''sequence``}.
In the following, we are going to evaluate the influence of the rules when they are successively added and exploited in the graph.
Note that rule \textit{''prior``} alone would merely return the initial values $\hat{S}^{DB\times Q}$.
Input to the graph are the pairwise descriptor similarities from the raw NetVLAD descriptors that serve as baseline as well (termed ''pairwise``).
All results are summarized in Table~\ref{tab:no_std_seq}.
\begin{table*}[tb]
\centering
\caption{Average precision for different configurations of the graph-based framework.
Colored arrows indicate large ($\ge$10\% better/ worse) or medium ($\ge$5\%) deviation compared to the configuration that exploits less prior knowledge (i.e., ''pairwise`` vs $\hat{S}^{DB}$; $\hat{S}^{DB}$ vs $\hat{S}^{DB}$+$\hat{S}^Q$; $\hat{S}^{DB}$+$\hat{S}^Q$ vs $\hat{S}^{DB}$+$\hat{S}^Q$+$\text{Seq}$). Bold text indicates the best performance per row and per intra-database source.}
\vspace{-0.3cm}
\label{tab:no_std_seq}

\resizebox{0.90\textwidth}{!}{%
\begin{tabular}{llll!{\VRule[2pt]}c||c|c|c||c|c|c}
    \hline
	 &   &                &                &                   &                                                                  \multicolumn{3}{c||}{$\hat{S}^{DB}$ from poses}                                                                   &                                                            \multicolumn{3}{c}{$\hat{S}^{DB}$ from descriptors}                                                             \\
	                       & &                  &                &                   &                   online                   &                                                      \multicolumn{2}{c||}{offline / delayed}                                                       &                   online                   &                                                   \multicolumn{2}{c}{offline / delayed}                                                    \\
	\textbf{Dataset}       &\textbf{Name}& \textbf{Database} & \textbf{Query} & \textbf{pairwise} &           $\hat{S}^{DB}_{pose}$            &         $\hat{S}^{DB}_{pose}{+}\hat{S}^Q_{desc}$         &                     $\hat{S}^{DB}_{pose}{+}\hat{S}^Q_{desc}{+}\text{Seq}$                     &           $\hat{S}^{DB}_{desc}$            &   $\hat{S}^{DB}_{desc}{+}\hat{S}^Q_{desc}$   &                       $\hat{S}^{DB}_{desc}{+}\hat{S}^Q_{desc}{+}\text{Seq}$                       \\ \hline
	Nordland               & Nordland\#1        & fall              & spring &       0.39        &     0.56 \color{supergood}{$\uparrow$}     &           0.56 \color{equal}{$\rightarrow$}            &             \textbf{0.95} \color{supergood}{$\uparrow$}              &     0.50 \color{supergood}{$\uparrow$}     &     0.52 \color{equal}{$\rightarrow$}      &               \textbf{0.93} \color{supergood}{$\uparrow$}                \\
	                       & Nordland\#2        & fall              & winter &       0.06        &     0.13 \color{supergood}{$\uparrow$}     &           0.17 \color{supergood}{$\uparrow$}           &             \textbf{0.49} \color{supergood}{$\uparrow$}              &     0.10 \color{supergood}{$\uparrow$}     &     0.14 \color{supergood}{$\uparrow$}     &               \textbf{0.42} \color{supergood}{$\uparrow$}                \\
	                       & Nordland\#3       & spring            & winter  &       0.11        &     0.19 \color{supergood}{$\uparrow$}     &           0.25 \color{supergood}{$\uparrow$}           &             \textbf{0.61} \color{supergood}{$\uparrow$}              &     0.16 \color{supergood}{$\uparrow$}     &     0.24 \color{supergood}{$\uparrow$}     &               \textbf{0.60} \color{supergood}{$\uparrow$}                \\
	                       & Nordland\#4       & winter            & spring  &       0.11        &     0.27 \color{supergood}{$\uparrow$}     &             0.29 \color{good}{$\nearrow$}              &             \textbf{0.71} \color{supergood}{$\uparrow$}              &     0.23 \color{supergood}{$\uparrow$}     &     0.24 \color{equal}{$\rightarrow$}      &               \textbf{0.59} \color{supergood}{$\uparrow$}                \\
	                       & Nordland\#5        & summer            & spring &       0.32        &     0.54 \color{supergood}{$\uparrow$}     &           0.54 \color{equal}{$\rightarrow$}            &             \textbf{0.95} \color{supergood}{$\uparrow$}              &     0.45 \color{supergood}{$\uparrow$}     &       0.48 \color{good}{$\nearrow$}        &               \textbf{0.92} \color{supergood}{$\uparrow$}                \\
	                       & Nordland\#6        & summer            & fall   &       0.63        &     0.80 \color{supergood}{$\uparrow$}     &           0.81 \color{equal}{$\rightarrow$}            &             \textbf{1.00} \color{supergood}{$\uparrow$}              &     0.74 \color{supergood}{$\uparrow$}     &     0.77 \color{equal}{$\rightarrow$}      &               \textbf{1.00} \color{supergood}{$\uparrow$}                \\
	StLucia                & StLucia\#1 & 100909$\_$0845    & 190809$\_$0845&       0.41        &     0.50 \color{supergood}{$\uparrow$}     &           0.51 \color{equal}{$\rightarrow$}            &             \textbf{0.75} \color{supergood}{$\uparrow$}              &     0.47 \color{supergood}{$\uparrow$}     &       0.50 \color{good}{$\nearrow$}        &               \textbf{0.74} \color{supergood}{$\uparrow$}                \\
	                       & StLucia\#2 & 100909$\_$1000    & 210809$\_$1000 &       0.47        &     0.55 \color{supergood}{$\uparrow$}     &           0.56 \color{equal}{$\rightarrow$}            &             \textbf{0.81} \color{supergood}{$\uparrow$}              &       0.52 \color{good}{$\nearrow$}        &       0.55 \color{good}{$\nearrow$}        &               \textbf{0.80} \color{supergood}{$\uparrow$}                \\
	                       & StLucia\#3 & 100909$\_$1210    & 210809$\_$1210&       0.51        &     0.59 \color{supergood}{$\uparrow$}     &           0.61 \color{equal}{$\rightarrow$}            &            \textbf{0.87} \color{supergood}{$\uparrow$}              &     0.56 \color{supergood}{$\uparrow$}     &       0.60 \color{good}{$\nearrow$}        &               \textbf{0.86} \color{supergood}{$\uparrow$}                \\
	                       & StLucia\#4 & 100909$\_$1410    & 190809$\_$1410&       0.38        &     0.50 \color{supergood}{$\uparrow$}     &           0.49 \color{equal}{$\rightarrow$}            &             \textbf{0.79} \color{supergood}{$\uparrow$}              &     0.46 \color{supergood}{$\uparrow$}     &       0.49 \color{good}{$\nearrow$}        &               \textbf{0.79} \color{supergood}{$\uparrow$}                \\
	                       & StLucia\#5 & 110909$\_$1545    & 180809$\_$1545&       0.27        &     0.37 \color{supergood}{$\uparrow$}     &           0.35 \color{equal}{$\rightarrow$}            &             \textbf{0.52} \color{supergood}{$\uparrow$}              &     0.32 \color{supergood}{$\uparrow$}     &       0.34 \color{good}{$\nearrow$}        &               \textbf{0.49} \color{supergood}{$\uparrow$}                \\
	CMU                    & CMU\#1      & 20110421          & 20100901 &       0.73        &       0.77 \color{good}{$\nearrow$}        &           0.79 \color{equal}{$\rightarrow$}            & \textbf{0.85} \color{good}{$\nearrow$}  &     0.74 \color{equal}{$\rightarrow$}      &     0.75 \color{equal}{$\rightarrow$}      &   \textbf{0.81} \color{good}{$\nearrow$}    \\
	                       & CMU\#2     & 20110421          & 20100915  &       0.77        &     0.81 \color{equal}{$\rightarrow$}      &           0.81 \color{equal}{$\rightarrow$}            & \textbf{0.87 \color{good}{$\nearrow$}}  &     0.78 \color{equal}{$\rightarrow$}      &     0.77 \color{equal}{$\rightarrow$}      & \textbf{0.85} \color{supergood}{$\uparrow$} \\
	                       & CMU\#3      & 20110421          & 20101221 &       0.56        &       0.59 \color{good}{$\nearrow$}        &           0.61 \color{equal}{$\rightarrow$}            & \textbf{0.66} \color{good}{$\nearrow$}  &     0.58 \color{equal}{$\rightarrow$}      &     0.59 \color{equal}{$\rightarrow$}      & \textbf{0.65} \color{supergood}{$\uparrow$} \\
	                       & CMU\#4     & 20110421          & 20110202  &       0.61        &     0.72 \color{supergood}{$\uparrow$}     &             0.77 \color{good}{$\nearrow$}              & \textbf{0.87} \color{supergood}{$\uparrow$} &       0.66 \color{good}{$\nearrow$}        &     0.69 \color{equal}{$\rightarrow$}      &               \textbf{0.83} \color{supergood}{$\uparrow$}                \\
	GardensPoint           & GP\#1  & day$\_$right      & day$\_$left  &       0.97        &     0.99 \color{equal}{$\rightarrow$}      &           0.99 \color{equal}{$\rightarrow$}            &                        \textbf{1.00} \color{equal}{$\rightarrow$}                         &     0.98 \color{equal}{$\rightarrow$}      &     0.98 \color{equal}{$\rightarrow$}      &                          \textbf{1.00} \color{equal}{$\rightarrow$}                           \\
	Walking                & GP\#2  & day$\_$right      & night$\_$right  &   0.51        &     0.57 \color{supergood}{$\uparrow$}     &             0.61 \color{good}{$\nearrow$}              &             \textbf{0.85} \color{supergood}{$\uparrow$}              &     0.56 \color{supergood}{$\uparrow$}     &       0.60 \color{good}{$\nearrow$}        &               \textbf{0.84} \color{supergood}{$\uparrow$}                \\
	                       & GP\#3 & day$\_$left       & night$\_$right &       0.40        &     0.45 \color{supergood}{$\uparrow$}     &             0.49 \color{good}{$\nearrow$}              &             \textbf{0.80} \color{supergood}{$\uparrow$}              &     0.44 \color{supergood}{$\uparrow$}     &     0.49 \color{supergood}{$\uparrow$}     &               \textbf{0.78} \color{supergood}{$\uparrow$}                \\
	Oxford                 & Oxford\#1       & 141209            & 141216  &       0.87        &   \textbf{0.92} \color{good}{$\nearrow$}   &           0.91 \color{equal}{$\rightarrow$}            &                  0.91 \color{equal}{$\rightarrow$}                   &     \textbf{0.90} \color{equal}{$\rightarrow$}      &      \textbf{0.90} \color{equal}{$\rightarrow$}      &                \textbf{0.90} \color{equal}{$\rightarrow$}                \\
	                       & Oxford\#2      & 141209            & 150203   &       0.93        & \textbf{0.96} \color{equal}{$\rightarrow$} &       \textbf{0.96} \color{equal}{$\rightarrow$}       &                        \textbf{0.96} \color{equal}{$\rightarrow$}                         & \textbf{0.95} \color{equal}{$\rightarrow$} & \textbf{0.95} \color{equal}{$\rightarrow$} &                          \textbf{0.95} \color{equal}{$\rightarrow$}                           \\
	                       & Oxford\#3      & 141209            & 150519   &       0.83        &     0.93 \color{supergood}{$\uparrow$}     &           0.95 \color{equal}{$\rightarrow$}            &                        \textbf{0.97} \color{equal}{$\rightarrow$}                         &       0.90 \color{good}{$\nearrow$}        &     0.93 \color{equal}{$\rightarrow$}      &                \textbf{0.95} \color{equal}{$\rightarrow$}                \\
	                       & Oxford\#4      & 150519            & 150203   &       0.85        &       0.92 \color{good}{$\nearrow$}        & 0.96 \color{equal}{$\rightarrow$} &              \textbf{0.97} \color{equal}{$\rightarrow$}              &     0.89 \color{equal}{$\rightarrow$}      &       0.94 \color{good}{$\nearrow$}        &                \textbf{0.96} \color{equal}{$\rightarrow$}                \\ \hline
\end{tabular}%
}
\vspace{-0.4cm}
\end{table*}

\subsubsection{Exploitation of intra-database similarities from poses or descriptors (rules \textit{''loop``} and \textit{''exclusion``})}\label{sec:exp2}
Intra-database similarities $\hat{S}^{DB}$ can be used in most place recognition setups as they can be acquired either from the pairwise image comparisons within the database or from poses, e.g., from GPS or SLAM.

Table~\ref{tab:no_std_seq} shows the results (indicated by $\hat{S}^{DB}_{pose}$ and $\hat{S}^{DB}_{desc}$) when intra-database similarities are used either from poses or descriptor comparisons.
In most cases, the pairwise performance is significantly improved and never gets worse.
If intra-database similarities from poses (here: GPS) are used, the performance gain is slightly better, presumably because place matchings and distinctions from poses are binary and less error prone.
However, even when the intra-database similarities from descriptor comparisons are used, the performance can be improved significantly.

The results support that most existing place recognition pipelines could be improved with the proposed graph-based framework together with intra-database similarities.
Moreover, since it is not necessary to know the query sequence in advance, the proposed graph-based framework can be employed in an online fashion.

\subsubsection{Exploitation of intra-database and intra-query similarities from poses or descriptors (rules \textit{''loop``} and \textit{''exclusion``})}
In addition to the previous experiment, supplementary intra-query similarities could be used to model dependencies within the graph not only between database images but also between query images.
We used intra-query similarities solely from descriptor comparisons since we do not assume global pose information during the query run; otherwise, place matchings could be received directly from pose-comparisons.

Table~\ref{tab:no_std_seq} shows the results (indicated by $\hat{S}^{DB}_{pose}{+}\hat{S}^Q_{desc}$ and $\hat{S}^{DB}_{desc}{+}\hat{S}^Q_{desc}$).
For intra-database similarities from poses, the performance is improved only for a few sequence-combinations compared to the performance with intra-database but without intra-query similarities.
When using intra-query similarities in addition to intra-database similarities from descriptors, the performance could be improved at least by additional $5\%$ for $50\%$ of all datasets.

The results indicate that additional data from intra-query similarities within the proposed graph-based framework can be used to improve the place recognition performance further.
It is important to note that the system again never performs worse in comparison to a graph with intra-database but without intra-query similarity exploitation.
The result is interesting, since intra-query similarities from descriptors could always be collected for a subsequent postprocessing.

\subsubsection{Additional exploitation of sequences within the graph (rules \textit{''loop``}, \textit{''exclusion``} and \textit{''sequence``})}
In this experiment, we used sequence information within the graph in addition to the intra-database and intra-query similarities.
Table~\ref{tab:no_std_seq} shows the results (indicated by $\hat{S}^{DB}_{pose}{+}\hat{S}^Q_{desc}{+}\text{Seq}$ and $\hat{S}^{DB}_{desc}{+}\hat{S}^Q_{desc}{+}\text{Seq}$).
With sequence information the graph can again significantly improve the place recognition performance compared to the previous experiments without this additional assumption no matter if the intra-database similarities come from poses or descriptor comparisons.
Moreover, the full setup of the graph with all proposed factors clearly outperforms the baseline.

\subsection{Combination with preprocessed descriptors}\label{sec:exp2_std}
The place recognition performance can be improved if the descriptors are preprocessed \cite{Schubert2020}.
To investigate the influence of preprocessing on all configurations of the graph from Sec.~\ref{sec:exp1}, we repeated all previous experiments with standardized image descriptors \cite{Schubert2020}.
Results are shown in Table~\ref{tab:std_seq} (left).
Since the preprocessing requires complete knowledge of the query descriptors, we do not provide results for the online configuration of the graph.
The full configuration of the graph with intra-set similarities and sequences can again show significant performance improvements for the majority of the sequence-combinations.
The results demonstrate that the graph-based framework benefits from better performing descriptors.
\begin{table*}[tb]
\centering
\caption{Evaluation of the average precision of the graph-based framework with pre- or postprocessing}
\vspace{-0.3cm}
\label{tab:std_seq}
\resizebox{1\textwidth}{!}{%
\begin{tabular}{l!{\VRule[2pt]}c||c|c||c|c!{\VRule[2pt]}c||c|c|c||c|c|c}
    \hline
 &                  &                       \multicolumn{2}{c||}{$\hat{S}^{DB}$ from poses}                &                                         \multicolumn{2}{c!{\VRule[2pt]}}{$\hat{S}^{DB}$ from descriptors}    & &                                         \multicolumn{3}{c||}{$\hat{S}^{DB}$ from poses}                &                                         \multicolumn{3}{c}{$\hat{S}^{DB}$ from descriptors} \\
	                    &                   &                                              \multicolumn{4}{c!{\VRule[2pt]}}{offline / delayed}      &                       &                     online&                                                             \multicolumn{2}{c||}{offline / delayed}     &online&\multicolumn{2}{c}{offline / delayed}                                                     \\
	\textbf{Dataset}    & \textbf{pairwise}            &   $\hat{S}^{DB}_{pose}{+}\hat{S}^Q_{desc}$   & $\hat{S}^{DB}_{pose}{+}\hat{S}^Q_{desc}{+}\text{Seq}$                   &         $\hat{S}^{DB}_{desc}{+}\hat{S}^Q_{desc}$         &                       $\hat{S}^{DB}_{desc}{+}\hat{S}^Q_{desc}{+}\text{Seq}$ & \textbf{pairwise} &$\hat{S}^{DB}_{pose}$            &   $\hat{S}^{DB}_{pose}{+}\hat{S}^Q_{desc}$   & $\hat{S}^{DB}_{pose}{+}\hat{S}^Q_{desc}{+}\text{Seq}$ &                 $\hat{S}^{DB}_{desc}$                  &         $\hat{S}^{DB}_{desc}{+}\hat{S}^Q_{desc}$         &                       $\hat{S}^{DB}_{desc}{+}\hat{S}^Q_{desc}{+}\text{Seq}$                      \\ \hline
	                    &                                                                                                                                   \multicolumn{5}{c!{\VRule[2pt]}}{\textbf{with preprocessing of descriptors} (Sec.~\ref{sec:exp2_std})}     & \multicolumn{7}{c}{\textbf{with postprocessing of similarities} (Sec.~\ref{sec:exp_postproc})}                                                                                                                                                                                                        \\ \hline
	Nordland\#1            &       0.61        &     0.62 \color{equal}{$\rightarrow$}      &    \textbf{0.99} \color{supergood}{$\uparrow$}    &           0.61 \color{equal}{$\rightarrow$}            &               \textbf{0.99} \color{supergood}{$\uparrow$}                &     0.95      &       \textbf{0.98} \color{equal}{$\rightarrow$}       &       \textbf{0.98} \color{equal}{$\rightarrow$}       &                0.96 \color{equal}{$\rightarrow$}                 & 0.97 \color{equal}{$\rightarrow$} & 0.97 \color{equal}{$\rightarrow$} &                \textbf{0.99} \color{equal}{$\rightarrow$}                \\
	  Nordland\#2                  &       0.26        &       0.27 \color{good}{$\nearrow$}        &    \textbf{0.85} \color{supergood}{$\uparrow$}    &           0.26 \color{equal}{$\rightarrow$}            &               \textbf{0.85} \color{supergood}{$\uparrow$}                &     0.63      &           0.85 \color{supergood}{$\uparrow$}           & \textbf{0.87} \color{equal}{$\rightarrow$} & 0.77 \color{superbad}{$\downarrow$} &           0.78 \color{supergood}{$\uparrow$}           & 0.81 \color{equal}{$\rightarrow$} & \textbf{0.90} \color{supergood}{$\uparrow$} \\
	 Nordland\#3                   &       0.37        &     0.38 \color{equal}{$\rightarrow$}      &    \textbf{0.88} \color{supergood}{$\uparrow$}    &           0.37 \color{equal}{$\rightarrow$}            &               \textbf{0.88} \color{supergood}{$\uparrow$}                &     0.80      &             0.86 \color{good}{$\nearrow$}              &         \textbf{0.93} \color{good}{$\nearrow$}         &     0.80 \color{superbad}{$\downarrow$}    &             0.86 \color{good}{$\nearrow$}              &         \textbf{0.92} \color{good}{$\nearrow$}         &                    0.91 \color{equal}{$\rightarrow$}                     \\
	 Nordland\#4                   &       0.37        &     0.38 \color{equal}{$\rightarrow$}      &    \textbf{0.88} \color{supergood}{$\uparrow$}    &           0.37 \color{equal}{$\rightarrow$}            &               \textbf{0.88} \color{supergood}{$\uparrow$}                &     0.79      &           0.92 \color{supergood}{$\uparrow$}           &       \textbf{0.94} \color{equal}{$\rightarrow$}       &    0.88 \color{bad}{$\searrow$}     &           0.91 \color{supergood}{$\uparrow$}           &       \textbf{0.92} \color{equal}{$\rightarrow$}       &                    0.90 \color{equal}{$\rightarrow$}                     \\
	 Nordland\#5                   &       0.58        &     0.60 \color{equal}{$\rightarrow$}      &    \textbf{0.98} \color{supergood}{$\uparrow$}    &           0.58 \color{equal}{$\rightarrow$}            &               \textbf{0.98} \color{supergood}{$\uparrow$}                &     0.95      &       \textbf{0.99} \color{equal}{$\rightarrow$}       &           0.98 \color{equal}{$\rightarrow$}            &                     0.97 \color{equal}{$\rightarrow$}                     &       \textbf{0.98} \color{equal}{$\rightarrow$}       &           0.97 \color{equal}{$\rightarrow$}            &                    \textbf{0.98} \color{equal}{$\rightarrow$}                     \\
	  Nordland\#6                  &       0.84        &     0.85 \color{equal}{$\rightarrow$}      &    \textbf{1.00} \color{supergood}{$\uparrow$}    &           0.84 \color{equal}{$\rightarrow$}            &               \textbf{1.00} \color{supergood}{$\uparrow$}                & \textbf{1.00} &       \textbf{1.00} \color{equal}{$\rightarrow$}       &       \textbf{1.00} \color{equal}{$\rightarrow$}       &                      \textbf{1.00} \color{equal}{$\rightarrow$}                       &       \textbf{1.00} \color{equal}{$\rightarrow$}       &       \textbf{1.00} \color{equal}{$\rightarrow$}       &                          \textbf{1.00} \color{equal}{$\rightarrow$}                           \\
	StLucia\#1             &       0.58        &     0.59 \color{equal}{$\rightarrow$}      &    \textbf{0.81} \color{supergood}{$\uparrow$}    &           0.58 \color{equal}{$\rightarrow$}            &               \textbf{0.80} \color{supergood}{$\uparrow$}                &     0.76      &             0.81 \color{good}{$\nearrow$}              &       \textbf{0.82} \color{equal}{$\rightarrow$}       &                0.79 \color{equal}{$\rightarrow$}                 &           0.79 \color{equal}{$\rightarrow$}            &       \textbf{0.82} \color{equal}{$\rightarrow$}       &                               0.81 \color{equal}{$\rightarrow$}                               \\
	 StLucia\#2                   &       0.61        &     0.62 \color{equal}{$\rightarrow$}      &    \textbf{0.84} \color{supergood}{$\uparrow$}    &           0.61 \color{equal}{$\rightarrow$}            &               \textbf{0.85} \color{supergood}{$\uparrow$}                &     0.84      & \textbf{0.85} \color{equal}{$\rightarrow$} & \textbf{0.85} \color{equal}{$\rightarrow$} &                0.84 \color{equal}{$\rightarrow$}                 &           0.84 \color{equal}{$\rightarrow$}            &       \textbf{0.85} \color{equal}{$\rightarrow$}       &                    0.84 \color{equal}{$\rightarrow$}                     \\
	StLucia\#3                    &       0.63        &     0.65 \color{equal}{$\rightarrow$}      &    \textbf{0.90} \color{supergood}{$\uparrow$}    &           0.64 \color{equal}{$\rightarrow$}            &               \textbf{0.90} \color{supergood}{$\uparrow$}                &     0.83      &           0.85 \color{equal}{$\rightarrow$}            &       \textbf{0.86} \color{equal}{$\rightarrow$}       &                0.84 \color{equal}{$\rightarrow$}                 &           0.84 \color{equal}{$\rightarrow$}            &       \textbf{0.86} \color{equal}{$\rightarrow$}       &                          \textbf{0.86} \color{equal}{$\rightarrow$}                           \\
	 StLucia\#4                   &       0.57        &     0.58 \color{equal}{$\rightarrow$}      &    \textbf{0.85} \color{supergood}{$\uparrow$}    &           0.57 \color{equal}{$\rightarrow$}            &               \textbf{0.85} \color{supergood}{$\uparrow$}                &     0.81      &         \textbf{0.87} \color{good}{$\nearrow$}         &           0.86 \color{equal}{$\rightarrow$}            &                0.86 \color{equal}{$\rightarrow$}                 &           0.85 \color{equal}{$\rightarrow$}            &           0.85 \color{equal}{$\rightarrow$}            &                          \textbf{0.86} \color{equal}{$\rightarrow$}                           \\
	 StLucia\#5                   &       0.44        &     0.45 \color{equal}{$\rightarrow$}      &    \textbf{0.68} \color{supergood}{$\uparrow$}    &           0.43 \color{equal}{$\rightarrow$}            &               \textbf{0.67} \color{supergood}{$\uparrow$}                &     0.55      &      \textbf{0.69} \color{supergood}{$\uparrow$}       &           0.66 \color{equal}{$\rightarrow$}            &        0.60 \color{bad}{$\searrow$}         &           0.63 \color{supergood}{$\uparrow$}           & 0.64 \color{equal}{$\rightarrow$} &     \textbf{0.71} \color{supergood}{$\uparrow$}      \\
	CMU\#1                 &       0.73        &       0.80 \color{good}{$\nearrow$}        &    \textbf{0.87} \color{good}{$\nearrow$}      &           0.76 \color{equal}{$\rightarrow$}            &   \textbf{0.83} \color{good}{$\nearrow$}    &     0.81      &       \textbf{0.85} \color{equal}{$\rightarrow$}       &       \textbf{0.85} \color{equal}{$\rightarrow$}       &                      \textbf{0.85} \color{equal}{$\rightarrow$}                       &           0.81 \color{equal}{$\rightarrow$}            & 0.82 \color{equal}{$\rightarrow$} &                \textbf{0.84} \color{equal}{$\rightarrow$}                \\
	 CMU\#2                   &       0.78        &       0.82 \color{good}{$\nearrow$}        &    \textbf{0.88} \color{good}{$\nearrow$}      &           0.79 \color{equal}{$\rightarrow$}            &   \textbf{0.86} \color{good}{$\nearrow$}    &     0.85      &       \textbf{0.87} \color{equal}{$\rightarrow$}       &       \textbf{0.87} \color{equal}{$\rightarrow$}       &                      \textbf{0.87} \color{equal}{$\rightarrow$}                       & 0.85 \color{equal}{$\rightarrow$} & 0.85 \color{equal}{$\rightarrow$} &                \textbf{0.86} \color{equal}{$\rightarrow$}                \\
	CMU\#3                    &       0.59        &     0.61 \color{equal}{$\rightarrow$}      &    \textbf{0.66} \color{good}{$\nearrow$}     &           0.59 \color{equal}{$\rightarrow$}            & \textbf{0.65} \color{supergood}{$\uparrow$} &     0.64      &           0.65 \color{equal}{$\rightarrow$}            &       \textbf{0.66} \color{equal}{$\rightarrow$}       &                      \textbf{0.66} \color{equal}{$\rightarrow$}                       &       \textbf{0.65} \color{equal}{$\rightarrow$}       &       \textbf{0.65} \color{equal}{$\rightarrow$}       &                          \textbf{0.65} \color{equal}{$\rightarrow$}                           \\
	CMU\#4                    &       0.71        &     0.79 \color{supergood}{$\uparrow$}     &      \textbf{0.89} \color{supergood}{$\uparrow$}        &           0.74 \color{equal}{$\rightarrow$}            &   \textbf{0.85} \color{supergood}{$\uparrow$}   &     0.75      &           0.84 \color{supergood}{$\uparrow$}           & \textbf{0.87} \color{equal}{$\rightarrow$} &            \textbf{0.87} \color{equal}{$\rightarrow$}            &             0.80 \color{good}{$\nearrow$}              &           0.81 \color{equal}{$\rightarrow$}            &                \textbf{0.84} \color{equal}{$\rightarrow$}                \\
	GP\#1        &       0.99        &     0.99 \color{equal}{$\rightarrow$}      &    \textbf{1.00} \color{equal}{$\rightarrow$}     &           0.99 \color{equal}{$\rightarrow$}            &                          \textbf{1.00} \color{equal}{$\rightarrow$}                           & \textbf{1.00} &       \textbf{1.00} \color{equal}{$\rightarrow$}       &       \textbf{1.00} \color{equal}{$\rightarrow$}       &                      \textbf{1.00} \color{equal}{$\rightarrow$}                       &       \textbf{1.00} \color{equal}{$\rightarrow$}       &       \textbf{1.00} \color{equal}{$\rightarrow$}       &                          \textbf{1.00} \color{equal}{$\rightarrow$}                           \\
	GP\#2             &       0.68        &     0.70 \color{equal}{$\rightarrow$}      &    \textbf{0.96} \color{supergood}{$\uparrow$}    &           0.69 \color{equal}{$\rightarrow$}            &               \textbf{0.95} \color{supergood}{$\uparrow$}                &     0.77      &           0.86 \color{supergood}{$\uparrow$}           & 0.89 \color{equal}{$\rightarrow$} &            \textbf{0.90} \color{equal}{$\rightarrow$}            &             0.84 \color{good}{$\nearrow$}              &           0.87 \color{equal}{$\rightarrow$}            & \textbf{0.98} \color{supergood}{$\uparrow$} \\
	GP\#3                    &       0.56        &     0.57 \color{equal}{$\rightarrow$}      &    \textbf{0.92} \color{supergood}{$\uparrow$}    &           0.57 \color{equal}{$\rightarrow$}            &               \textbf{0.91} \color{supergood}{$\uparrow$}                &     0.76      &           0.84 \color{supergood}{$\uparrow$}           & 0.87 \color{equal}{$\rightarrow$} &            \textbf{0.89} \color{equal}{$\rightarrow$}            &             0.80 \color{good}{$\nearrow$}              &             0.85 \color{good}{$\nearrow$}              & \textbf{0.96} \color{supergood}{$\uparrow$} \\
	Oxford\#1              &       0.92        & \textbf{0.93} \color{equal}{$\rightarrow$} &    0.92 \color{equal}{$\rightarrow$}     & \textbf{0.92} \color{equal}{$\rightarrow$} &                \textbf{0.92} \color{equal}{$\rightarrow$}                &     0.89      &       \textbf{0.92} \color{equal}{$\rightarrow$}       &           0.90 \color{equal}{$\rightarrow$}            &                     0.90 \color{equal}{$\rightarrow$}                     &       \textbf{0.92} \color{equal}{$\rightarrow$}       &           0.90 \color{equal}{$\rightarrow$}            &                         0.91 \color{equal}{$\rightarrow$}                         \\
	Oxford\#2                    &       0.96        &     0.97 \color{equal}{$\rightarrow$}      &    \textbf{0.98} \color{equal}{$\rightarrow$}     &       \textbf{0.97} \color{equal}{$\rightarrow$}       &                          \textbf{0.97} \color{equal}{$\rightarrow$}                           &     0.92      &       \textbf{0.95} \color{equal}{$\rightarrow$}       &       \textbf{0.95} \color{equal}{$\rightarrow$}       &                      \textbf{0.95} \color{equal}{$\rightarrow$}                       & 0.94 \color{equal}{$\rightarrow$} & 0.94 \color{equal}{$\rightarrow$} &                \textbf{0.96} \color{equal}{$\rightarrow$}                \\
	Oxford\#3                    &       0.91        &       0.97 \color{good}{$\nearrow$}        &    \textbf{0.98} \color{equal}{$\rightarrow$}     &           0.95 \color{equal}{$\rightarrow$}            &                          \textbf{0.97} \color{equal}{$\rightarrow$}                           &     0.88      &   \textbf{0.96} \color{good}{$\nearrow$}   & \textbf{0.96} \color{equal}{$\rightarrow$} &            \textbf{0.96} \color{equal}{$\rightarrow$}            &             0.94 \color{good}{$\nearrow$}              & 0.95 \color{equal}{$\rightarrow$} &                \textbf{0.97} \color{equal}{$\rightarrow$}                \\
	Oxford\#4                    &       0.94        &     0.97 \color{equal}{$\rightarrow$}      &    \textbf{0.98} \color{equal}{$\rightarrow$}     &           0.96 \color{equal}{$\rightarrow$}            &                \textbf{0.98} \color{equal}{$\rightarrow$}                &     0.93      &           0.96 \color{equal}{$\rightarrow$}            &       \textbf{0.97} \color{equal}{$\rightarrow$}       &                      \textbf{0.97} \color{equal}{$\rightarrow$}                       &           0.95 \color{equal}{$\rightarrow$}            &           0.96 \color{equal}{$\rightarrow$}            &                \textbf{0.98} \color{equal}{$\rightarrow$}                \\\hline
\end{tabular}%
}
\vspace{-0.3cm}
\end{table*}

\subsection{Combination with sequence-based postprocessing}\label{sec:exp_postproc}
In the next experiment, a modified version of SeqSLAM~\cite{seqSLAM} without local contrast normalization and the single matching constraint is used; otherwise SeqSLAM would fail on datasets with in-sequence loops.
It postprocesses the inter-set similarities $S^{DB\times Q}$ either from the pairwise descriptor comparison (baseline) or from all configurations of the graph from Sec.~\ref{sec:exp1}.

Results are shown in Table~\ref{tab:std_seq} (right);
note that descriptor comparisons from the raw NetVLAD descriptors \textit{without} sequence-based postprocessing are used as inputs to the graphs.
The baseline performance after postprocessing compared to the baseline performance without postprocessing (Table~\ref{tab:no_std_seq}) is already comparatively high.
Accordingly, it is hard to achieve high performance improvements.
Nonetheless, the graph-based framework could improve the results in almost $50\%$ of the cases often by more than $10\%$ -- however, intra-database similarities from both poses and descriptors seem to be sufficient, since more information could not be used for further performance improvements on most datasets.

\subsection{Combination with pre- and postprocessing}
We also conducted experiments with preprocessing from Sec.~\ref{sec:exp2_std} \textit{and} postprocessing from Sec.~\ref{sec:exp_postproc}.
The baseline performance got already almost perfect for most of the datasets, so that performances could only be improved slightly by less than $5\%$ with the graph-based approach.
Again, the performance was never worse than the baseline.

\subsection{Comparison with state-of-the-art sequence-based methods}\label{sec:seq}
To compare performance and runtime of our method with approaches from the literature, we conduced additional experiments with the sequence-based methods SeqSLAM~\cite{seqSLAM}, MCN~\cite{Neubert2019}, VPR~\cite{Vysotska2017} and ABLE~\cite{able}.
The sequence length was $L=11$, if required.
The experiments were performed twice without and with feature standardization~\cite{Schubert2020} for descriptor preprocessing.

Table~\ref{tab:seq} shows the achieved performances.
Our method clearly outperformed the compared approaches and achieved the best performance on most datasets.
Only on Nordland SeqSLAM and ABLE achieved better performance, since they benefit from constant camera speed in database and query.
With feature standardization, ABLE could additionally achieve best performance on GardensPoint, and MCN performed best on three Oxford datasets.

Runtimes were measured on an Intel i7-7700K CPU with 64GB RAM.
The maximum runtimes per query for all methods are shown in Table~\ref{tab:seq} (bottom).
Our method required approx. 5.7sec per query on Oxford\#2 with 3413 database images, while all other approaches needed less than 500msec.
All reported runtimes are presumably sufficient for applications like loop closure detection in SLAM, since loops need not to be detected with the full frame rate of a camera.
Sec.~\ref{sec:discussion} provides some more discussion on the computational efficiency and potential improvements.

\boldmath
\begin{table*}[tb]
    \centering
    \caption{Average precision and maximum runtime per query of our method compared to sequence-based approaches from the literature. Colored arrows indicate deviation from ''pairwise``. Bold text indicates the best performance per row and per preprocessing.}
    \vspace{-0.3cm}
    \label{tab:seq}
    \resizebox{0.88\textwidth}{!}{%
        \begin{tabular}{l!{\VRule[2pt]}P{0.6cm}|P{1.8cm}|P{1.2cm}|P{1.0cm}|P{1.0cm}|P{1.0cm}!{\VRule[2pt]}P{0.6cm}|P{1.8cm}|P{1.2cm}|P{1.0cm}|P{1.0cm}|P{1.0cm}}
            \hline
            \textbf{Dataset}    & \textbf{pair- wise}& \textbf{$\hat{S}^{DB}_{desc}{+}\hat{S}^Q_{desc}\allowbreak{+}\text{Seq}$ (ours)} &       \textbf{SeqSLAM} \cite{seqSLAM}       &    \textbf{MCN} \cite{Neubert2019}     &  \textbf{VPR} \cite{Vysotska2017}   &          \textbf{ABLE} \cite{able}& \textbf{pair- wise}& \textbf{$\hat{S}^{DB}_{desc}{+}\hat{S}^Q_{desc}\allowbreak{+}\text{Seq}$ (ours)} &       \textbf{SeqSLAM} \cite{seqSLAM}       &    \textbf{MCN} \cite{Neubert2019}     &  \textbf{VPR} \cite{Vysotska2017}   &          \textbf{ABLE} \cite{able}          \\ \hline
            
            &  \multicolumn{6}{c!{\VRule[2pt]}}{\textbf{(online) w/o preprocessing}} &  \multicolumn{6}{c}{\textbf{(offline) descriptor preprocessing with feature standardization \cite{Schubert2020}}} \\\hline
Nordland\#1 & 0.39 \color{equal}{} &0.93 \color{supergood}{$\uparrow$} &0.89 \color{supergood}{$\uparrow$} &0.52 \color{supergood}{$\uparrow$} &0.68 \color{supergood}{$\uparrow$} &\textbf{0.98} \color{supergood}{$\uparrow$} &0.61 \color{equal}{} &\textbf{0.99} \color{supergood}{$\uparrow$} &0.92 \color{supergood}{$\uparrow$} &0.73 \color{supergood}{$\uparrow$} &0.36 \color{superbad}{$\downarrow$} &\textbf{0.99} \color{supergood}{$\uparrow$} \\
Nordland\#2 & 0.06 \color{equal}{} &0.42 \color{supergood}{$\uparrow$} &0.72 \color{supergood}{$\uparrow$} &0.21 \color{supergood}{$\uparrow$} &0.26 \color{supergood}{$\uparrow$} &\textbf{0.74} \color{supergood}{$\uparrow$} &0.26 \color{equal}{} &0.85 \color{supergood}{$\uparrow$} &0.77 \color{supergood}{$\uparrow$} &0.33 \color{supergood}{$\uparrow$} &0.33 \color{supergood}{$\uparrow$} &\textbf{0.90} \color{supergood}{$\uparrow$} \\
Nordland\#3 & 0.11 \color{equal}{} &0.60 \color{supergood}{$\uparrow$} &0.83 \color{supergood}{$\uparrow$} &0.24 \color{supergood}{$\uparrow$} &0.39 \color{supergood}{$\uparrow$} &\textbf{0.85} \color{supergood}{$\uparrow$} &0.37 \color{equal}{} &0.88 \color{supergood}{$\uparrow$} &0.85 \color{supergood}{$\uparrow$} &0.44 \color{supergood}{$\uparrow$} &0.34 \color{bad}{$\searrow$} &\textbf{0.96} \color{supergood}{$\uparrow$} \\
Nordland\#4 & 0.11 \color{equal}{} &0.59 \color{supergood}{$\uparrow$} &0.81 \color{supergood}{$\uparrow$} &0.29 \color{supergood}{$\uparrow$} &0.32 \color{supergood}{$\uparrow$} &\textbf{0.85} \color{supergood}{$\uparrow$} &0.37 \color{equal}{} &0.88 \color{supergood}{$\uparrow$} &0.86 \color{supergood}{$\uparrow$} &0.44 \color{supergood}{$\uparrow$} &0.34 \color{bad}{$\searrow$} &\textbf{0.96} \color{supergood}{$\uparrow$} \\
Nordland\#5 & 0.32 \color{equal}{} &0.92 \color{supergood}{$\uparrow$} &0.91 \color{supergood}{$\uparrow$} &0.41 \color{supergood}{$\uparrow$} &0.64 \color{supergood}{$\uparrow$} &\textbf{0.97} \color{supergood}{$\uparrow$} &0.58 \color{equal}{} &0.98 \color{supergood}{$\uparrow$} &0.92 \color{supergood}{$\uparrow$} &0.71 \color{supergood}{$\uparrow$} &0.36 \color{superbad}{$\downarrow$} &\textbf{0.99} \color{supergood}{$\uparrow$} \\
Nordland\#6 & 0.63 \color{equal}{} &\textbf{1.00} \color{supergood}{$\uparrow$} &0.95 \color{supergood}{$\uparrow$} &0.51 \color{superbad}{$\downarrow$} &0.89 \color{supergood}{$\uparrow$} &\textbf{1.00} \color{supergood}{$\uparrow$} &0.84 \color{equal}{} &\textbf{1.00} \color{supergood}{$\uparrow$} &0.95 \color{supergood}{$\uparrow$} &0.91 \color{good}{$\nearrow$} &0.37 \color{superbad}{$\downarrow$} &\textbf{1.00} \color{supergood}{$\uparrow$} \\
StLucia\#1 & 0.41 \color{equal}{} &\textbf{0.74} \color{supergood}{$\uparrow$} &0.12 \color{superbad}{$\downarrow$} &0.56 \color{supergood}{$\uparrow$} &0.47 \color{supergood}{$\uparrow$} &0.39 \color{equal}{$\rightarrow$} &0.58 \color{equal}{} &\textbf{0.80} \color{supergood}{$\uparrow$} &0.12 \color{superbad}{$\downarrow$} &0.57 \color{equal}{$\rightarrow$} &0.01 \color{superbad}{$\downarrow$} &0.45 \color{superbad}{$\downarrow$} \\
StLucia\#2 & 0.47 \color{equal}{} &\textbf{0.80} \color{supergood}{$\uparrow$} &0.12 \color{superbad}{$\downarrow$} &0.61 \color{supergood}{$\uparrow$} &0.48 \color{equal}{$\rightarrow$} &0.46 \color{equal}{$\rightarrow$} &0.61 \color{equal}{} &\textbf{0.85} \color{supergood}{$\uparrow$} &0.12 \color{superbad}{$\downarrow$} &0.62 \color{equal}{$\rightarrow$} &0.00 \color{superbad}{$\downarrow$} &0.50 \color{superbad}{$\downarrow$} \\
StLucia\#3 & 0.51 \color{equal}{} &\textbf{0.86} \color{supergood}{$\uparrow$} &0.14 \color{superbad}{$\downarrow$} &0.65 \color{supergood}{$\uparrow$} &0.47 \color{bad}{$\searrow$} &0.47 \color{bad}{$\searrow$} &0.63 \color{equal}{} &\textbf{0.90} \color{supergood}{$\uparrow$} &0.13 \color{superbad}{$\downarrow$} &0.62 \color{equal}{$\rightarrow$} &0.01 \color{superbad}{$\downarrow$} &0.49 \color{superbad}{$\downarrow$} \\
StLucia\#4 & 0.38 \color{equal}{} &\textbf{0.79} \color{supergood}{$\uparrow$} &0.14 \color{superbad}{$\downarrow$} &0.53 \color{supergood}{$\uparrow$} &0.42 \color{supergood}{$\uparrow$} &0.50 \color{supergood}{$\uparrow$} &0.57 \color{equal}{} &\textbf{0.85} \color{supergood}{$\uparrow$} &0.14 \color{superbad}{$\downarrow$} &0.55 \color{equal}{$\rightarrow$} &0.03 \color{superbad}{$\downarrow$} &0.55 \color{equal}{$\rightarrow$} \\
StLucia\#5 & 0.27 \color{equal}{} &\textbf{0.49} \color{supergood}{$\uparrow$} &0.13 \color{superbad}{$\downarrow$} &0.40 \color{supergood}{$\uparrow$} &0.41 \color{supergood}{$\uparrow$} &0.32 \color{supergood}{$\uparrow$} &0.44 \color{equal}{} &\textbf{0.67} \color{supergood}{$\uparrow$} &0.13 \color{superbad}{$\downarrow$} &0.51 \color{supergood}{$\uparrow$} &0.00 \color{superbad}{$\downarrow$} &0.44 \color{equal}{$\rightarrow$} \\
CMU\#1 & 0.73 \color{equal}{} &\textbf{0.81} \color{supergood}{$\uparrow$} &0.03 \color{superbad}{$\downarrow$} &\textbf{0.81} \color{supergood}{$\uparrow$} &0.47 \color{superbad}{$\downarrow$} &0.71 \color{equal}{$\rightarrow$} &0.73 \color{equal}{} &\textbf{0.83} \color{supergood}{$\uparrow$} &0.03 \color{superbad}{$\downarrow$} &0.82 \color{supergood}{$\uparrow$} &0.00 \color{superbad}{$\downarrow$} &0.72 \color{equal}{$\rightarrow$} \\
CMU\#2 & 0.77 \color{equal}{} &\textbf{0.85} \color{supergood}{$\uparrow$} &0.07 \color{superbad}{$\downarrow$} &0.79 \color{equal}{$\rightarrow$} &0.47 \color{superbad}{$\downarrow$} &0.71 \color{bad}{$\searrow$} &0.78 \color{equal}{} &\textbf{0.86} \color{supergood}{$\uparrow$} &0.07 \color{superbad}{$\downarrow$} &0.79 \color{equal}{$\rightarrow$} &0.06 \color{superbad}{$\downarrow$} &0.71 \color{bad}{$\searrow$} \\
CMU\#3 & 0.56 \color{equal}{} &\textbf{0.65} \color{supergood}{$\uparrow$} &0.05 \color{superbad}{$\downarrow$} &0.60 \color{good}{$\nearrow$} &0.43 \color{superbad}{$\downarrow$} &0.53 \color{bad}{$\searrow$} &0.59 \color{equal}{} &\textbf{0.65} \color{supergood}{$\uparrow$} &0.06 \color{superbad}{$\downarrow$} &0.61 \color{equal}{$\rightarrow$} &0.09 \color{superbad}{$\downarrow$} &0.53 \color{superbad}{$\downarrow$} \\
CMU\#4 & 0.61 \color{equal}{} &\textbf{0.83} \color{supergood}{$\uparrow$} &0.13 \color{superbad}{$\downarrow$} &0.82 \color{supergood}{$\uparrow$} &0.47 \color{superbad}{$\downarrow$} &0.49 \color{superbad}{$\downarrow$} &0.71 \color{equal}{} &\textbf{0.85} \color{supergood}{$\uparrow$} &0.13 \color{superbad}{$\downarrow$} &0.82 \color{supergood}{$\uparrow$} &0.00 \color{superbad}{$\downarrow$} &0.55 \color{superbad}{$\downarrow$} \\
GP\#1 & 0.97 \color{equal}{} &\textbf{1.00} \color{equal}{$\rightarrow$} &0.42 \color{superbad}{$\downarrow$} &0.98 \color{equal}{$\rightarrow$} &0.69 \color{superbad}{$\downarrow$} &\textbf{1.00} \color{equal}{$\rightarrow$} &0.99 \color{equal}{} &\textbf{1.00} \color{equal}{$\rightarrow$} &0.43 \color{superbad}{$\downarrow$} &0.99 \color{equal}{$\rightarrow$} &0.67 \color{superbad}{$\downarrow$} &\textbf{1.00} \color{equal}{$\rightarrow$} \\
GP\#2 & 0.51 \color{equal}{} &\textbf{0.84} \color{supergood}{$\uparrow$} &0.29 \color{superbad}{$\downarrow$} &0.62 \color{supergood}{$\uparrow$} &0.47 \color{bad}{$\searrow$} &0.79 \color{supergood}{$\uparrow$} &0.68 \color{equal}{} &0.95 \color{supergood}{$\uparrow$} &0.34 \color{superbad}{$\downarrow$} &0.70 \color{equal}{$\rightarrow$} &0.67 \color{equal}{$\rightarrow$} &\textbf{0.97} \color{supergood}{$\uparrow$} \\
GP\#3 & 0.40 \color{equal}{} &\textbf{0.78} \color{supergood}{$\uparrow$} &0.14 \color{superbad}{$\downarrow$} &0.44 \color{good}{$\nearrow$} &0.29 \color{superbad}{$\downarrow$} &0.77 \color{supergood}{$\uparrow$} &0.56 \color{equal}{} &0.91 \color{supergood}{$\uparrow$} &0.16 \color{superbad}{$\downarrow$} &0.60 \color{good}{$\nearrow$} &0.67 \color{supergood}{$\uparrow$} &\textbf{0.94} \color{supergood}{$\uparrow$} \\
Oxford\#1 & 0.87 \color{equal}{} &\textbf{0.90} \color{equal}{$\rightarrow$} &0.05 \color{superbad}{$\downarrow$} &0.87 \color{equal}{$\rightarrow$} &0.51 \color{superbad}{$\downarrow$} &0.76 \color{superbad}{$\downarrow$} &0.92 \color{equal}{} &0.92 \color{equal}{$\rightarrow$} &0.05 \color{superbad}{$\downarrow$} &\textbf{0.97} \color{good}{$\nearrow$} &0.00 \color{superbad}{$\downarrow$} &0.81 \color{superbad}{$\downarrow$} \\
Oxford\#2 & 0.93 \color{equal}{} &\textbf{0.95} \color{equal}{$\rightarrow$} &0.03 \color{superbad}{$\downarrow$} &0.91 \color{equal}{$\rightarrow$} &0.52 \color{superbad}{$\downarrow$} &0.82 \color{superbad}{$\downarrow$} &0.96 \color{equal}{} &0.97 \color{equal}{$\rightarrow$} &0.02 \color{superbad}{$\downarrow$} &\textbf{0.98} \color{equal}{$\rightarrow$} &0.00 \color{superbad}{$\downarrow$} &0.86 \color{superbad}{$\downarrow$} \\
Oxford\#3 & 0.83 \color{equal}{} &\textbf{0.95} \color{supergood}{$\uparrow$} &0.05 \color{superbad}{$\downarrow$} &0.94 \color{supergood}{$\uparrow$} &0.54 \color{superbad}{$\downarrow$} &0.66 \color{superbad}{$\downarrow$} &0.91 \color{equal}{} &0.97 \color{good}{$\nearrow$} &0.04 \color{superbad}{$\downarrow$} &\textbf{0.99} \color{good}{$\nearrow$} &0.00 \color{superbad}{$\downarrow$} &0.76 \color{superbad}{$\downarrow$} \\
Oxford\#4 & 0.85 \color{equal}{} &\textbf{0.96} \color{supergood}{$\uparrow$} &0.02 \color{superbad}{$\downarrow$} &0.86 \color{equal}{$\rightarrow$} &0.45 \color{superbad}{$\downarrow$} &0.82 \color{equal}{$\rightarrow$} &0.94 \color{equal}{} &\textbf{0.98} \color{equal}{$\rightarrow$} &0.02 \color{superbad}{$\downarrow$} &0.97 \color{equal}{$\rightarrow$} &0.01 \color{superbad}{$\downarrow$} &0.90 \color{equal}{$\rightarrow$} \\\hline
           max. runtime per&\multirow{2}{*}{-}&\multirow{2}{*}{5.6sec}&\multirow{2}{*}{24msec}&\multirow{2}{*}{280msec}&\multirow{2}{*}{3.7msec}&\multirow{2}{*}{63$\mu$sec}&\multirow{2}{*}{-}&\multirow{2}{*}{5.7sec} &\multirow{2}{*}{23msec}&\multirow{2}{*}{441msec}&\multirow{2}{*}{3.6msec}&\multirow{2}{*}{68$\mu$sec}\\
           query (Oxford\#2)&&&&&&&&&&&\\\hline

        \end{tabular}
    }
\vspace{-0.3cm}
\end{table*}
\unboldmath

\section{DISCUSSION AND CONCLUSION}\label{sec:discussion}
The previous sections presented our approach to use a graphical model as a flexible framework to model different kinds of additional information available in place recognition.
The experiments demonstrated that the presented method can considerably improve place recognition results in various configurations in terms of available data (e.g., poses of database images), the subset of applied rules (e.g., using sequences or not), or restriction to online place recognition.
The representation and optimization using graphical models offers a high degree of flexibility.
In the remainder of this last section, we want to discuss aspects of the proposed system and some particularly interesting possible extensions.

The graph-based optimization is performed on the pairwise descriptor similarities $S^{DB\times Q}$.
This makes it relatively independent of the actually chosen image descriptor, which may require only a slight parameter adjustment.
This property even allows an optimization of place descriptor similarities from different sensor modalities like LiDAR.

In this paper, we defined several factor-types to express prior knowledge (``rules'') about place recognition problems.
Each factor implements a cost function for a rule. 
Often, there are alternative formulations of a cost function for a rule.
For example, the cost function in case of no loop in query ($\hat{s}^Q_{ij}\downarrow$; Eq.~(\ref{eq:cost_dissim_q})) or database ($\hat{s}^{DB}_{ij}\downarrow$; Eq.~(\ref{eq:cost_dissim_db})) is defined as multiplication $(s_1\cdot s_2)^2$.
Especially for this rule, alternative formulations may apply like $\min(s_1, s_2)^2$ or $\max(s_1+s_2-1, 0)^2$; both are piecewise linear which could be beneficial.

Factor graphs are often (but not exclusively) used in combination with probabilities. Presumably, a more probabilistic view on the proposed graph-based framework could provide additional insights.
For instance, the chosen cost functions (\ref{eq:cost_unary}), (\ref{eq:cost_sim_db}) and (\ref{eq:cost_dissim_q}) with structure $(s_1-s_2)^2$ can be considered as the negative log-likelihood of a single Gaussian:
\vspace{-0.02cm}
\begin{align}
 (s_1-s_2)^2 \Leftrightarrow -\ln(e^{-(s_1-s_2)^2})
\end{align}
Moreover, a piecewise linear formulation of the discussed factor above could help to formulate the proposed graph-based framework in a more probabilistic way; for instance a cost function $\min(s_1, s_2)^2$ corresponds to the negative log-likelihood of a maximum of two Gaussians:
\begin{align}
 \min(s_1, s_2)^2 \Leftrightarrow -\ln(\max(e^{-(s_1-0)^2}, e^{-(s_2-0)^2})
\end{align}

These cost functions, however, solely work on the descriptor similarities.
In the related work (Sec.~\ref{sec:related_work}), we already mentioned the important difference to pose graph SLAM, which solely works on spatial poses (and not their similarities).
An interesting question for future work is whether the proposed approach can be extended to also directly work on descriptors instead of their similarities (i.e., the variables would be descriptor vectors, not their scalar similarities).
This significantly increases the complexity of the optimization problem, but could allow the simultaneous optimization of spatial poses and descriptors for a potentially tightly coupled loop closure detection and SLAM.

Even without such an extension, as indicated in Sec.~\ref{sec:algo} and Sec.~\ref{sec:optim}, the number of factors or connections between nodes can get quite huge, and grows cubically if intra-set similarities are used.
In our experiments, we addressed the problem by dividing $S^{DB\times Q}$ into patches if both intra-set similarities were used (Sec.~\ref{sec:implementation}).
Presumably, using more efficient implementations, e.g. C++ implementations in Ceres \cite{ceres}, can improve memory consumption and computational efficiency.
Another promising direction are approximation techniques like a systematic removal of low-relevance connections in the graph.
Finally, different optimization techniques could be used: In earlier work on graph optimization, minimization techniques based on hill-climbing algorithms like ICM (iterated conditional modes) were used \cite[p.599]{Koller2009}; these may allow a different and more compact representation and optimization of the graph.


\end{document}